\RequirePackage{fix-cm}
\documentclass[smallextended]{svjour3}       
\smartqed  
\usepackage{graphicx}
\usepackage{ctable}
\usepackage{multirow}
\usepackage{color}
\usepackage{longtable}
\usepackage{subfigure}
\usepackage{fancyhdr}

\usepackage{amssymb}
\begin{document}

\title{Self-supervised Learning via Cluster Distance Prediction for Operating Room Context Awareness
}

\titlerunning{Self-Supervised Operating Room Awareness}        

\author{Idris Hamoud \and
Alexandros Karargyris \and
Aidean Sharghi \and
Omid Mohareri \and
Nicolas Padoy 
}


\institute{Idris Hamoud \and  Nicolas Padoy \at
              ICube, University of Strasbourg, CNRS, France \\
              \email{\{ihamoud, npadoy\}@unistra.fr}
           \and
           Alexandros Karargyris \and Nicolas Padoy  \at
              IHU Strasbourg, France \\
              \email{\{alexandros.karargyris, nicolas.padoy\}@ihu-strasbourg.eu}
           \and
           Aidean Sharghi \and Omid Mohareri  \at
              Intuitive Surgical Inc., Sunnyvale, USA \\
              \email{\{Aidean.SharghiKarganroodi, Omid.Mohareri\}@intusurg.com}
}


\maketitle

\begin{abstract} \quad

\paragraph{Purpose}Semantic segmentation and activity classification are key components to create intelligent surgical systems able to understand and assist clinical workflow. In the Operating Room, semantic segmentation is at the core of creating robots aware of clinical surroundings, whereas activity classification aims at understanding OR workflow at a higher level. State-of-the-art semantic segmentation and activity recognition approaches are fully supervised, which is not scalable. Self-supervision, can decrease the amount of annotated data needed.
\paragraph{Methods}We propose a new 3D self-supervised task for OR scene understanding utilizing OR scene images captured with ToF cameras. Contrary to other self-supervised approaches, where handcrafted pretext tasks are focused on 2D image features, our proposed task consists of predicting relative 3D distance of image patches by exploiting the depth maps. By learning 3D spatial context, it generates discriminative features for our downstream tasks.

\paragraph{Results}Our approach is evaluated on two tasks and datasets containing multi-view data captured from clinical scenarios. We demonstrate a noteworthy improvement of performance on both tasks, specifically on low-regime data where utility of self-supervised learning is the highest.
\paragraph{Conclusion}We propose a novel privacy-preserving self-supervised approach utilizing depth maps. Our proposed method shows performance on par with other self-supervised approaches, and could be an interesting way to alleviate the burden of full-supervision.

\end{abstract}

\section{Introduction}

Robotic surgery is applied across a wide range of common general surgical procedures. It has been demonstrated that robot-assisted surgery could lead to a better overall operative outcome in certain procedures \cite{Sheetz2020}. Nevertheless, it is also responsible for an increase in Operating Room (OR) workflow complexity because of additional actions and checkups that are required. This amplified complexity in the intraoperative routine can potentially make caretakers more prone to errors \cite{Catchpole2015SafetyEA}.

In spite of that, high-tech surgical devices offer the possibility of gathering information from multiple sensors. This additional information can then be processed through a data-driven workflow analysis system to create a context-aware operating room as emphasized by \emph{Dias et al.} \cite{Dias}. This information is crucial for understanding surgical operations and could encourage reducing the number of surgical errors by improving communication and coordination between staff members.
Many studies \cite{LowReso,MVORVinkle,Luo2018VisionBasedDA,Sharghi2020,Li2020AR3,Ressucit2013,DBLP:journals/corr/abs-1811-12296} have been conducted using computer vision and deep learning algorithms to generate a 3D understanding of the scenes in the OR. These approaches require significant human labor from experts to manually annotate the collected data. 

In our proposed method we aim to develop a self-supervised approach to reduce the need for annotations in data-driven methods. We show the value and generalizability of our approach by conducting data-efficiency experiments on two separate datasets and two different tasks: activity classification and semantic segmentation. 
To address privacy concerns and preserve the anonymity of the staff and patients involved in the videos, we only use depth information acquired from ToF cameras. Furthermore, we also benchmark other self-supervised methods \cite{Oord2018,RotNet2018} using our depth images. These two methods were originally designed and evaluated using RGB natural images.


In this paper, we introduce a novel 3D based pretext task that exploits the depth maps capturing the operating room.
Our method is a two-step approach. We first learn the relative euclidean distance between superpixel regions of the image using an encoder-decoder architecture as  pretraining. As the distance between objects is invariant to viewpoints and the training dataset contains images from multiple viewpoints, viewpoint invariance is then naturally embedded in the feature space. Once the pretext task is learned, we then initialize the encoder for the downstream task with our pretrained weights and finetune the network using progressive amounts of labels.

The contributions of this paper are two fold: (1) we introduce a new self-supervised pretext task based on depth information which exploits the spatial information provided by ToF cameras; (2)  we benchmark the proposed method with two other self-supervised methods, RotNet \cite{RotNet2018} and CPC v2 \cite{Oord2018}, on two different datasets \cite{Sharghi2020,Li2020AR3} for two different tasks related to surgical activity monitoring, demonstrating the effectiveness of our method.

\section{Related work}
\begin{figure}
\centering
\includegraphics[width=0.90\textwidth]{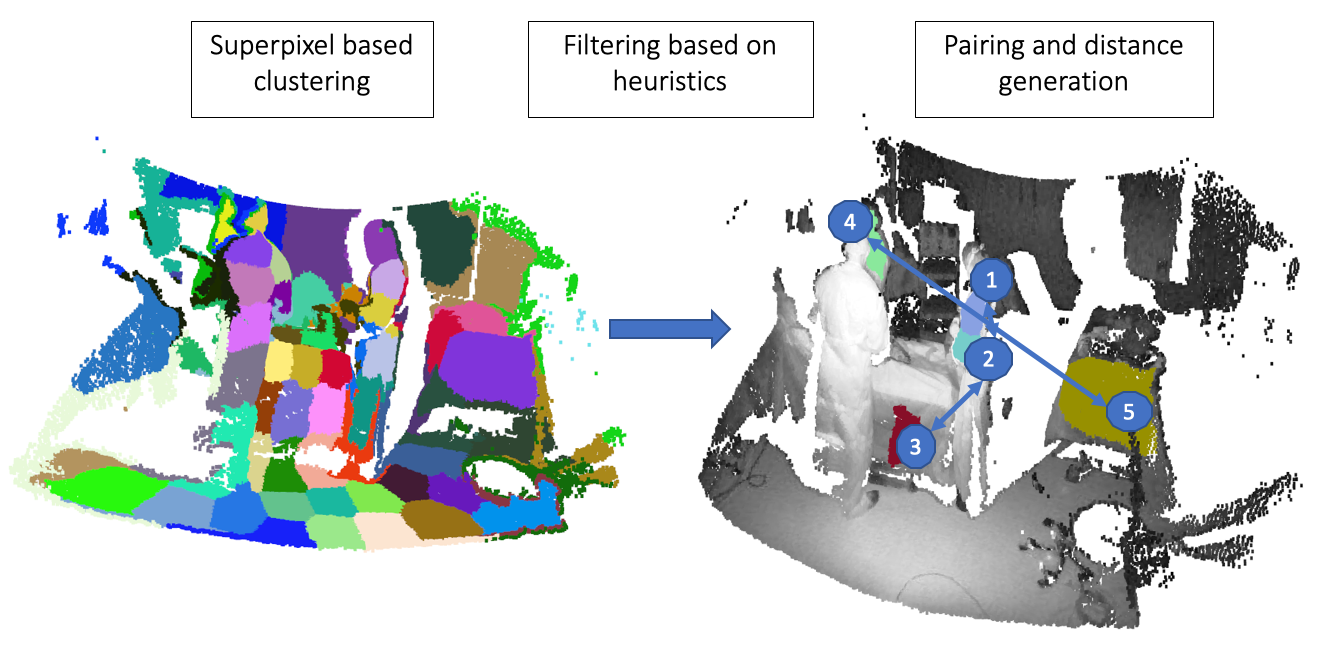}
\caption{Pretext task annotation generation process using SLIC \cite{SLIC2012} superpixel segmentation.} 
\label{fig1:imgs}
\end{figure}

\subsection{Operating Room Context Awareness}
Surgical workflow monitoring has gained a lot of attention recently. Many studies have been proposed to develop artificial intelligence (AI) based context aware systems in the OR to help scale up computer assisted operations. 

In general, most work has focused on images capturing the surgeon's point of view \cite{Twinanda2017,AlHajj2018}.
A significant amount of work have focused on endoscopic cholecystectomy procedures \cite{Twinanda2017}, making the most of the publicly available Cholec80 dataset \cite{Twinanda2017}. 
The CATARACTS \cite{AlHajj2018} dataset is another popular dataset containing annotations for both cataract surgery tool segmentation and phase prediction. The videos are acquired from two non-overlapping views, one from the microscope and the other one from a camera capturing the tray.

Our work, in contrast, is more in line with other approaches that focus on a more holistic view of the OR.
In \cite{Twinanda2017MultiStreamDA,Kadkhodamohammadi17}, \emph{Twinanda et al.} and \emph{Kadkhodamohammadi et al.} initiated the field by introducing new datasets and methods for both human pose estimation and action classification on RGB+D data. 
In \cite{MVORVinkle}, authors introduce MVOR, a new dataset containing annotated images from a hybrid OR during real clinical interventions. In \cite{LowReso}, they proposed to address the problem of human pose estimation in the OR using only low resolution privacy preserving images with a super-resolution block to increase the feature map resolution.  
In \cite{DBLP:journals/corr/abs-1811-12296}, the authors provide a comparison of state-of-the-art face detectors on OR data and also present an approach to train a face detector for the OR by exploiting non-annotated images. Similarly, authors on \cite{Ressucit2013} propose a vision system using Markov Logic Networks to segment activities in trauma resuscitation videos recorded from one ceiling camera.

Possibly the most relevant works to our contribution are \emph{Li et al.}, \emph{Sharghi et al.} and \emph{Schmidt et al.} \cite{Li2020AR3,Schmidt2021,Sharghi2020}. In these papers, authors provide new datasets for OR environment awareness during {\it da Vinci} based robotic procedures. \emph{Li et al.} \cite{Li2020AR3} propose a multiview fusion framework using point cloud information to improve semantic segmentation. \emph{Sharghi et al.} \cite{Sharghi2020} propose a new temporal model for long video understanding and phase detection, while \emph{Schmidt et al.} \cite{Schmidt2021} propose a new attention layer to merge information from multiple views in the OR.

\subsection{Self-supervised learning }
In computer vision, most self-supervised approaches are evaluated on image-level classification. Previously designed pretext tasks such as \cite{NorooziF16,RotNet2018,DoerschGE15} do not embed pixel-level information and can consequently underperform on dense predictive downstream tasks such as semantic segmentation.

In \emph{Doersch et al.} the authors proposed a new self-supervised task, which consists of predicting the relative position of square patches on an RGB image. Similarly, \emph{Noroozi et al.} also introduced a patch-based task, where patches of the images are permuted and a convolutional neural network is used to learn the applied permutation. In \emph{Gidaris et al.}, four different rotations are applied to the image and the CNN learns the applied rotations.

More recently, alternative approaches based on contrastive learning \cite{MOCO,SimCLR,BYOL,Barlow}, and clustering methods \cite{SwAV,DINO} have been proposed. These methods focus on creating negative and positive samples using data augmentations. The embeddings from the positive samples are pulled together, while those from negative samples are pulled apart. 
In CPC v2 \cite{Oord2018}, the authors came up with a different perspective. They propose a patch-based approach, in which the image is split into a grid of overlapping patches. Those patches are encoded through a feature extractor to gather their feature vector representation. The feature vectors from patches above a certain level are used to linearly predict the feature vectors of those below this same level.

Extensive studies of these contrastive self-supervised approaches have been presented for medical images in \cite{Taleb,azizi2021big}. In \emph{Taleb et al.} \cite{Taleb}, 3D voxel versions of five self-supervised approaches have been proposed and evaluated on semantic segmentation of CT and MR scans for both pancreatic and brain tumor segmentation. For surgical videos, \emph{Ross et al.} propose a recolorization pretext task using a GAN-like architecture for endoscopic image semantic segmentation \cite{Ro2018ExploitingTP}.
Lately more dense contrastive methods have been proposed to embed more region-specific feature learning \cite{DenseCLR,Ouyang2020}. For instance, \emph{Ouyang et al.} \cite{Ouyang2020} introduce a superpixel based contrastive loss to increase performance on CT-scan organ segmentation. Our work differs in that we are designing a task for a different modality, namely depth images, which effectively incorporates 2.5D spatial information.

\section{Methodology}
\subsection{Proposed Pretext Task}
The robotic operating room is a highly streamlined platform in which the persons and objects are expected to follow a certain protocol and be at a specific place at a specific time. 
Relative positions of objects in the room can provide powerful information  to integrate in surgical workflow analysis. In contrast to other pretext tasks such as \emph{Doersch et al.}\cite{DoerschGE15}, where only the 2D relative position of patches is used, our pretext task aims at taking advantage of the depth information by predicting relative distance of objects in 3D without any annotations. 
To this end we propose a new sampling method to extract homogeneous clusters from our depth map (see Fig. \ref{fig1:imgs}), so that each cluster belongs to one specific object. Depth maps from different views are analyzed independently in our work. Inspired by \emph{Ouyang et al.}, we propose a superpixel based approach to compute our clusters. Superpixels tend to be small-scale, dense image regions that offer a nice and smooth unsupervised segmentation \cite{graphcutSP,SLIC2012}. In this work, we employed the SLIC method \cite{SLIC2012}, because it is faster and more memory efficient compared to other existing methods. In addition to these quantifiable benefits, SLIC is easy to use, offers flexibility in the compactness and number of the superpixels it generates. 

\subsubsection{Self-supervised labeling strategy}
To extract homogeneous regions from our depth map, SLIC \cite{SLIC2012} is used to define regions of an overall identical depth and thus likely belonging to the same object (see Fig. \ref{fig4:all_model_arch}). In our experiments, we used the scikit-image implementation.
The approximate number of segments to be generated by SLIC was chosen as 500, based on experiments and qualitative analysis on the used training data. We also choose $\sigma = 3$ as width of the Gaussian smoothing kernel for preprocessing the image to take into account the noise in the depth maps. The compactness is chosen as  $\kappa = 3$ to balance importance between spatial proximity in the image coordinates and  depth proximity in the image intensity. 
Once the superpixels are generated, we filter them using heuristics based on the convexity of the superpixel region (solidity over 0.75), the disparity of the depth inside the cluster (std under 0.2m), and the number of missing values in the cluster (less than 5\%) to prevent having regions with too much noise. These heuristics were chosen based on preliminary studies  on our training data. We are only interested in compact regions with very low deviation in the three spatial directions. These superpixels can then be mapped to a set of points in the corresponding point cloud. The available camera intrinsics and the depth information is used to compute the corresponding coordinates for each point in our clusters.
In the end, we obtain for each image $I$ a set of point clusters $\{SP_1, ..., SP_N\}$ and define the distance between two clusters as the euclidean distance between the two centroids of the two point clouds (see Fig. \ref{fig4:all_model_arch}): \begin{equation}
  SP_1 = \{x_1^1,..., x_{N_1}^1\}   \;\;\;  SP_2 = \{x_1^2,..., x_{N_2}^2\}   \;\;\;   x_i^j \in \mathbb{R}^3 
\end{equation}
\begin{equation}
  C_1 = \frac{\sum_{i=1}^{N_1} x_i^1}{N_1} \;\;\;   C_2 = \frac{\sum_{i=1}^{N_2} x_i^2}{N_2}
\end{equation}
\begin{equation}
  SP_{dist} = \|C_1-C_2\|_2  \;\;\;.
\end{equation}
The euclidean distance between the superpixel clusters is regressed on the distance between the corresponding learnt representations in the feature space, as expressed below:
\begin{equation}
h_{SP_1} = f_{extract}(D_{SP_1})   \;\;\;  h_{SP_2} = f_{extract}(D_{SP_2}) 
\end{equation}
\begin{equation}
l_2 = \|h_{SP_1}-h_{SP_2}\|_2   
\end{equation}
\begin{equation}
L_{pretext} = \|l_2-SP_{dist}\|_1  \;\;\;. 
\end{equation}
where $h_{SP_i}$ represents the feature vector extracted from our network $f_{extract}$ by giving the corresponding depth map patch $D_{SP_i}$ as the input. $L_{pretext}$ is the loss regressed by our network illustrated in Fig. \ref{fig4:all_model_arch} and described in the next section.

\subsubsection{Encoder-Decoder architecture}
\paragraph{Feature Extraction:} ResNet-50 \cite{He2016_V2} has been successfully employed in many works for both semantic segmentation and activity detection \cite{DBLP:journals/corr/abs-1812-00033}. In this work, we also utilize the same architecture as our backbone visual feature extraction model. This model maps $224 \times 224 \times 1$ depth maps to a feature space of size {$9 \times 9 \times 2048$. It is trained on frames extracted from the videos, without any temporal context.

\paragraph{Architecture:} To be able to retrieve features at a superpixel level, we need to upscale our feature space. In this work, we use FCN-32 \cite{LongSD14} with a ResNet backbone \cite{He2016_V2} as our encoder-decoder architecture for the sake of simplicity and computational efficiency. This architecture only contains one single deconvolutional layer followed by one upsampling module, resulting in a $224 \times 224 \times 32$ feature map.
For our semantic segmentation experiments we keep the same architecture as for our pretext task, whereas for our activity classification experiments, we only keep the pretrained ResNet encoder with a global average pooling and a fully connected layer on top.

\paragraph{Superpixel Sampling module (SPS):} We sample the cluster features from the decoder output map. We use the external knowledge provided by the superpixel map to retrieve the position from where we sample our features. For each pair of clusters, we consider the smallest bounding box around the superpixel and extract the features from those two bounding boxes (see Fig. \ref{fig4:all_model_arch}).
Once those features are pulled from the decoder output, we resize them to compute an element-wise $l_{2}$ loss between them. The superpixel features are resized to $20 \times 20 \times 32$. 

\begin{figure}
\centering
\includegraphics[width=0.90\textwidth]{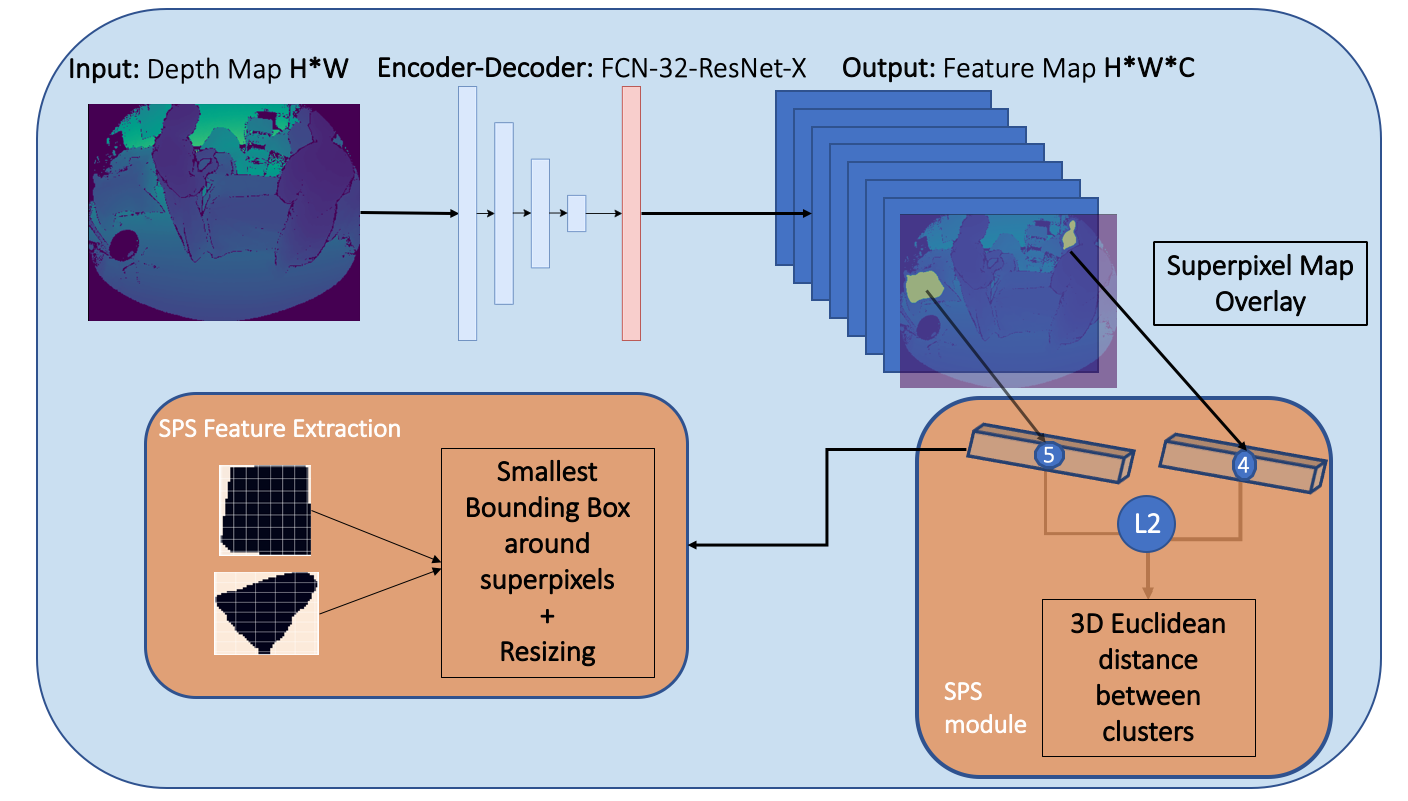}
\caption{Framework used for self-supervised learning; numbers 4 and 5 on feature vectors refer to Figure \ref{fig1:imgs}.}\label{fig4:all_model_arch}
\end{figure}

\subsection{Semi-supervised Learning: Semantic Segmentation \& Activity Classification}\label{section:dataset}

\paragraph{Methodology:}
It has been demonstrated by some critical studies on self-supervised learning  that results were dependent of the complexity of the dataset, of the downstream task at hand, of the architecture used and of course of the amount of supervision \cite{Newell2020HowUI,Asano2020ACA}.
Our aim is to answer those concerns by demonstrating the utility of our method on two different architectures with different complexities, and two different downstream tasks. 

%


%

%
%

\paragraph{Evaluation Metrics:} We use the same evaluation metrics as in the related publications  \cite{Li2020AR3,Schmidt2021,Sharghi2020}, where mean average precision (mAP) and mean intersection over union (mIoU) are used to effectively compare the results. 

\section{Experiments and Results}
\subsection{Operating Room Awareness Datasets}

To demonstrate the applicability of our method, we use the  following two recent datasets captured from the OR and containing depth image data.  

\begin{figure}
\centering
{\includegraphics[width=1.0\textwidth]{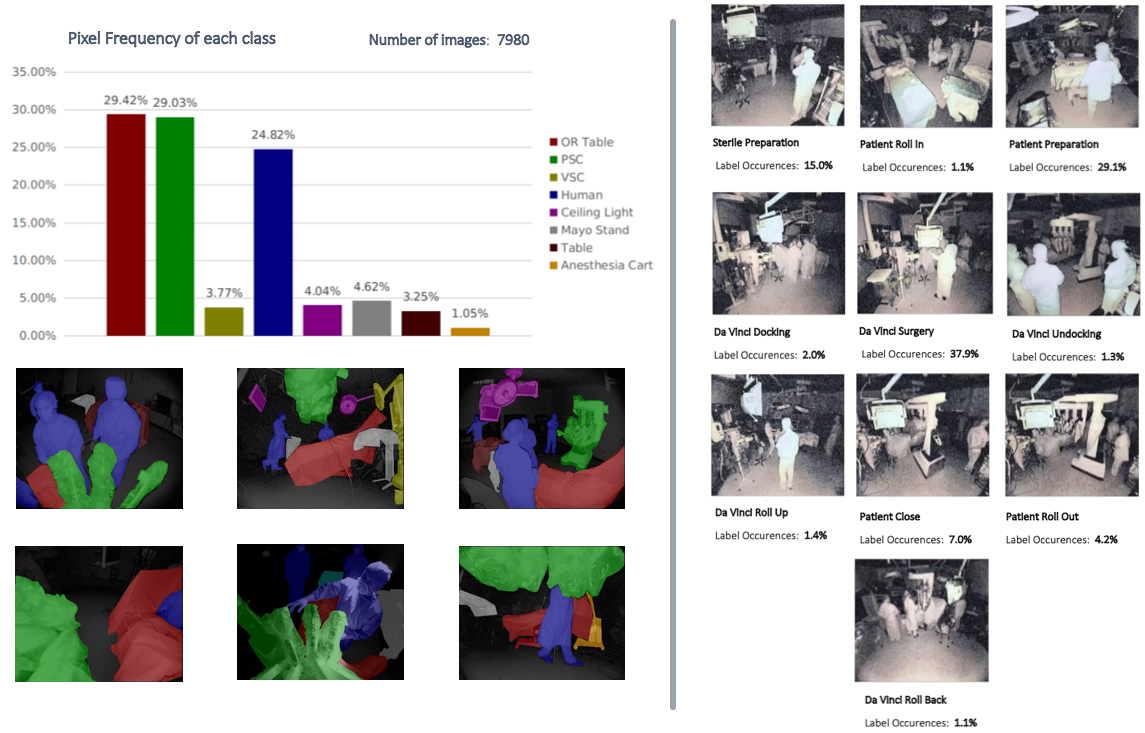}}
\caption{Overview of the semantic segmentation and activity classification datasets \cite{Li2020AR3,Sharghi2020}.} \label{fig2:semsegOverview}
\end{figure}

{\bfseries Semantic Segmentation Dataset}
This dataset introduced by \cite{Li2020AR3} contains 7980 densely annotated depth images captured by four different ToF cameras placed on the da Vinci robot. The data has been collected in a clinical development lab, where different robot-assisted laparoscopic procedures are simulated, and videos are taken by the ToF cameras. The cameras were attached on the patient side cart (PSC) and the vision side cart (VSC). Eight different objects are annotated in this dataset. Please refer to Fig. \ref{fig2:semsegOverview} for more details on the represented classes.

{\bfseries Activity Classification Dataset}
This dataset was introduced by \cite{Sharghi2020,Schmidt2021}. Videos were recorded from a single medical facility with two robotic ORs and da Vinci Xi systems. Two imaging carts were placed in each room, each equipped with two ToF cameras. In total 400 videos long of around two hours were collected. The operations are conducted by 16 different surgeons and contain different type of procedures.
Each of those videos are annotated using ten clinically relevant OR activities (see Fig. \ref{fig2:semsegOverview}:}).  
We use a subset of the dataset containing 34 cases with 28 robot-assisted surgical procedures and 6 non robot-assisted procedures. We sample the videos at 1 frame per second resulting in a dataset containing 98929 images, a subset large enough to showcase the usefulness of our method. We split our data into training, validation and testing so that there is no overlap between cases.
The two previously proposed methods \cite{Sharghi2020,Schmidt2021} on this dataset focused on activity segmentation using spatio-temporal models. In our case, for the sake of demonstrating the benefits of our self-supervised approach, the designed downstream task is frame wise activity classification.

%

\subsection{Unsupervised evaluation of self-supervised task}
\begin{figure}
\centering
{\includegraphics[width=0.70\textwidth]{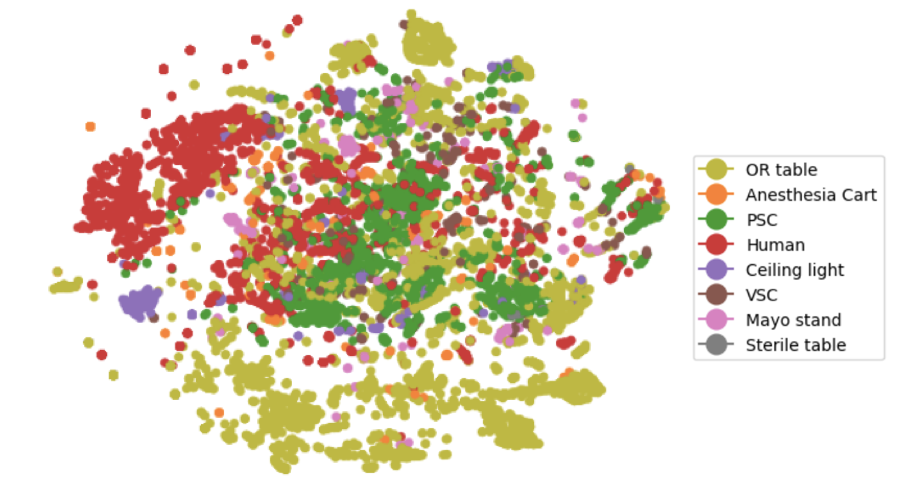}}
\caption{t-SNE \cite{JMLR:v9:vandermaaten08a}
visualization of the superpixel features learnt from the pretext task.} \label{fig5:tSNE}
\end{figure}
The t-SNE \cite{JMLR:v9:vandermaaten08a} method is a dimensionality reduction method widely used in the computer vision field to evaluate qualitatively the features learnt by a neural network.
In our case, each point in the 2D point-cloud represents the features belonging to one superpixel and extracted from our SPS module. 

As superpixel segmentation provides an oversegmentation of our image, it is very relevant to cluster the features extracted from those regions.
These features are obtained without any supervision, and as we can see on Fig. \ref{fig5:tSNE}, our pretext task manages to learn identifiable clusters for most of the semantic classes appearing in our semantic segmentation dataset. This is visible for  less frequent classes such as ceiling light and also for more frequent classes like human and OR table.

\subsection{Semi-supervised learning and data efficiency experiments}
\paragraph{Pretraining protocol:}
We evaluate our method against three baselines:

(1) In the first setting, we train from scratch without any pretraining to measure the benefits of the other self-supervised methods.

(2) The first self-supervised baseline is RotNet \cite{RotNet2018}, which is trained for 200 epochs to predict different rotations that have been applied to the initial image following the implementation from \emph{Gidaris et al.} \cite{RotNet2018}.

(3) The second self-supervised baseline is CPC v2 \cite{Oord2018} trained using the authors' implementation for 200 epochs.

Regarding our pretext task, it is trained with a learning rate of  3e-4 and a batch size of 32 for 200 epochs. We also make sure that all our baselines are trained fairly by saving only the best performing model on our validation dataset over the 200 epochs.

\paragraph{Finetuning protocol:}
We evaluate our method in a semi-supervised manner with different amounts of annotated data (2\%, 5\%, 10\%, 20\%, 50\%, 100\%). We follow the usual semi-supervised learning protocol and run our experiments with both ResNet-18 and ResNet-50 to show that our results are not dependent on the network complexity. Our results are averaged across five different random splits for all different data regimes, to account for the randomness introduced by sampling small amount of data, as done in \cite{Ro2018ExploitingTP}.

\begin{figure}
\centering
\includegraphics[width=1.0\textwidth]{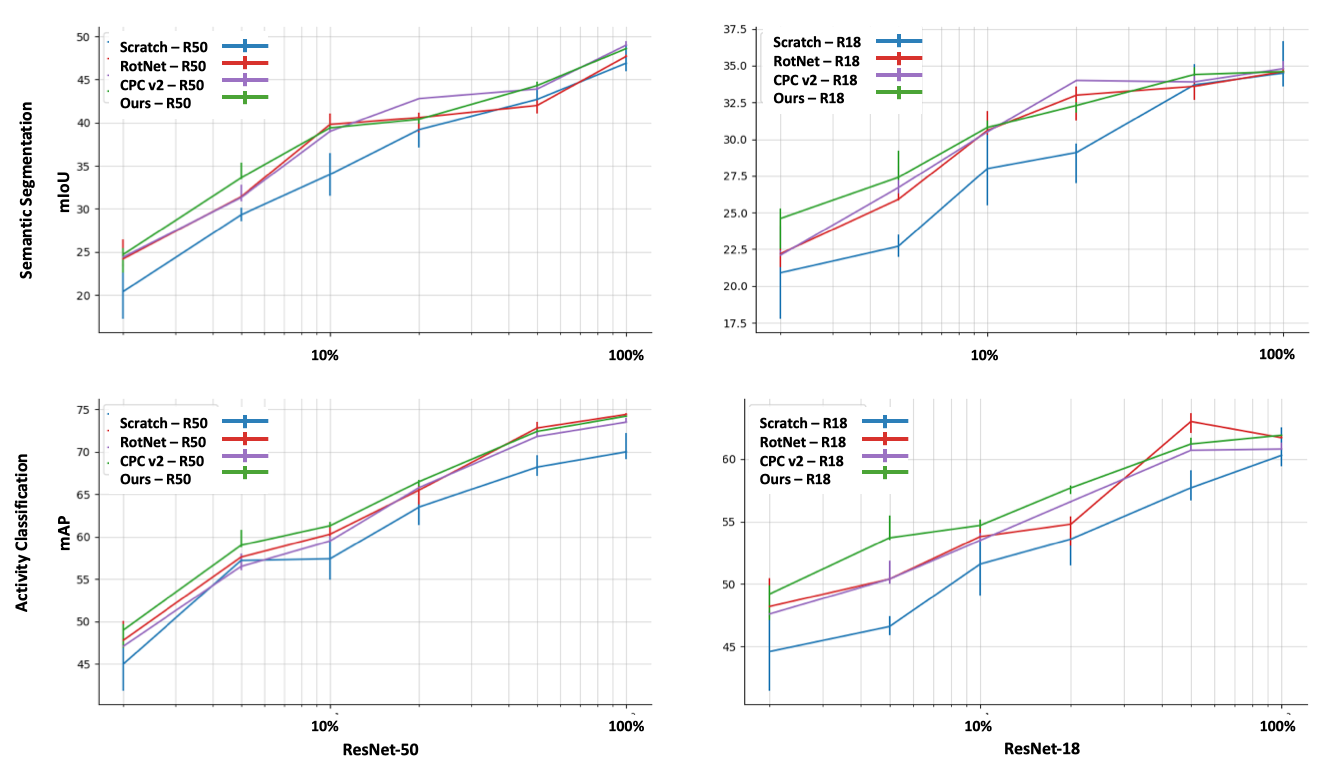}
\caption{Median mIoU and mAP with Interquartile Range (IQR) as a
function of training available labels as described section 4.3. Our method outperforms the baseline method without pretraining and is on par with other self-supervised methods.} 
\label{fig6:results}
\end{figure}
%

\paragraph{Downstream performance:}
The results are shown in Fig. \ref{fig6:results}. They demonstrate the usefulness of the proposed task as a new pretraining task, as it outperforms training from scratch as expected. The gap becomes smaller for all self-supervised pretraining experiments as we gradually increase the amount of supervision.  
Our pretrained task also outperforms the two other self-supervised approaches on low-data regime consistently across the two architectures and the two tasks. It  performs similarly on high-data regime for both tasks. 
On activity classification, for example, our proposed task gets the same mAP performance as the training from scratch using only half the amount of annotations at 5$\%$ and 50$\%$.

\paragraph{Statistical Significance Analysis:}

We further measure the statistical significance of our proposed pretext task performance compared to the "scratch" baseline using a Wilcoxon signed-rank test. We perform the significance analysis for all data fractions, based upon the collected $p$-values that were adjusted by Dunnett's test across splits and by Bonferroni-Holm correction across data fractions. Our proposed method shows significant ($p$ $<<$ 0.05) improvement on both downstream tasks for the low regime data, up to 20\% for semantic segmentation and up to 10\% for activity classification. It even shows contributions significant enough to beat a narrower 0.01 significant level on the three lower data fractions (2\%, 5\%, 10\%) for semantic segmentation.

\section{Conclusion and Future Work}
In this paper, we investigate the benefits of using self-supervision on depth maps for surgical operating room segmentation and recognition tasks.
We propose a new 3D based pretext task that exploits 3D information from the OR layout. We then compare our method with other self-supervised approaches like RotNet \cite{RotNet2018} and CPC v2 \cite{Oord2018}. Our experiments show noteworthy gains in performance in low-regime data compared to the baseline methods, thus highlighting the usefulness of the proposed pretext task.

{\bfseries Acknowledgements} This work is supported by a PhD fellowship from Intuitive Surgical and by French state funds managed within the “Plan Investissements d’Avenir” by the ANR (reference ANR-10-IAHU-02). 

{\bfseries Informed consent} Data has been collected within an Institutional Review Board (IRB) approved study and all participant's informed consent have been obtained. 

{\bfseries Ethical approval} All procedures performed in studies involving human participants were in accordance with the ethical standards of the institutional and/or national research committee and with the 1964 Helsinki declaration and its later amendments or comparable ethical standards.
%


%

\bibliographystyle{spmpsci}      
\bibliography{reference}
\end{document}